\documentclass[runningheads]{llncs}
\usepackage{graphicx}
\usepackage{amsmath,amssymb} 
\usepackage{color}

\usepackage[width=122mm,left=12mm,paperwidth=146mm,height=193mm,top=12mm,paperheight=217mm]{geometry}

\usepackage{wrapfig}
\usepackage[hang]{subfigure}
\usepackage{amsmath}
\usepackage{amssymb}
\usepackage{stmaryrd}
\usepackage{algorithm}
\usepackage{algorithmic}
\usepackage{tabularx}
\usepackage{multirow}
\usepackage{array}
\usepackage{enumerate}
\usepackage{paralist}
\usepackage{tabularx}

\newcommand\etal{\emph{et.al.}} 
\newcommand\ie{\emph{i.e.}} 
\newcommand\cf{\emph{c.f.}} 
\newcommand\etc{\emph{etc.}}
 
\newcommand\eg{\emph{e.g.}}

\begin{document}
\title{Open-World Stereo Video Matching \\ with Deep RNN}

\titlerunning{Open-World Stereo Video Matching with Deep RNN}
%
\author{Yiran Zhong\inst{1,3,4}, Hongdong Li\inst{1,3}, Yuchao Dai\inst{2}}
%
\authorrunning{Y. Zhong, H. Li and Y. Dai}
%
\institute{Australian National University, Australia \and
Northwestern Polytechnical University, China \\ \and
Australian Centre for Robotic Vision, Australia \\ \and
Data61 CSIRO, Australia \\
\email{\{yiran.zhong,hongdong.li\}@anu.edu.au \
daiyuchao@nwpu.edu.cn}}

%
\maketitle              
\begin{abstract}
Deep Learning based stereo matching methods have shown great successes and achieved top scores across different benchmarks. However, like most data-driven methods, existing deep stereo matching networks suffer from some well-known drawbacks such as requiring large amount of labeled training data, and that their performances are fundamentally limited by the generalization ability.  In this paper, we propose a novel Recurrent Neural Network (RNN) that takes a continuous (possibly previously unseen) stereo video as input, and directly predicts a depth-map at each frame without a pre-training process, and without the need of ground-truth depth-maps as supervision. Thanks to the recurrent nature (provided by two convolutional-LSTM blocks), our network is able to memorize and learn from its past experiences, and modify its inner parameters (network weights) to adapt to previously unseen or unfamiliar environments. This suggests a remarkable generalization ability of the net, making it applicable in an {\em open world} setting. Our method works robustly with changes in scene content, image statistics, and lighting and season conditions {\em etc}.  By extensive experiments, we demonstrate that the proposed method seamlessly adapts between different scenarios.  Equally important, in terms of the stereo matching accuracy, it outperforms state-of-the-art deep stereo approaches on standard benchmark datasets such as KITTI and Middlebury stereo.

\end{abstract}

\section{Introduction}
Stereo matching is a classic problem in computer vision, and it has been extensively studied in the literature for decades. Recently, deep learning based stereo matching methods are taking over, becoming one of the best performing approaches. As an evidence, they occupy the leader-boards for almost all the standard stereo matching benchmarks (\eg, KITTI\cite{Geiger2012CVPR}, Middlebury stereo \cite{Scharstein2002}).

However, there exists a considerable gap between the success of these ``deep stereo matching methods'' on somewhat artificially created benchmark datasets and their real-world performances when being employed ``in the wild'' (open world), probably for the following reasons:
\begin{enumerate}[(1)]
\item Most of the existing deep stereo matching methods are supervised learning based methods, for which the training process demands massive annotated training samples.  In the context of stereo matching, getting large amount of training data (\ie ground-truth disparity/depth maps) is an extremely expensive task.
\item The performance of existing deep stereo matching methods and their applicability in real-world scenarios are fundamentally limited by their generalization ability:  like most data-driven methods, they only work well on testing data that are sufficiently similar to the training data. Take autonomous driving for example, a deep stereo matching network trained in one city, under one traffic condition, might not work well in another city, under different lighting conditions.
\item So far, most deep stereo matching methods exclusively focus on processing single pair of stereo images in a frame-by-frame manner, while in real world stereo camera captures continuous video. The rich temporal information contained in the stereo video has not been exploited to improve the stereo matching performance or robustness.
\end{enumerate}

In this paper, we tackle all the above drawbacks with current deep stereo matching methods. We propose a novel deep Recurrent Neural Network (RNN) that computes a depth/disparity map continuously from stereo video, without any pre-training process. Contrary to conventional stereo matching methods (\eg, \cite{Hirschmuller2008,KendallMDHKBB17}) which focus on processing a single pair of stereo images individually, this work is capitalized on explicitly exploiting the temporally dynamic nature of stereo video input.

Our deep stereo video matching network, termed as ``OpenStereoNet'' is not fixed, but changes its inner parameters continuously as long as new stereo frames being fed into the network. This enables our network to adapt to changing situations (\eg changing lighting condition, changing image contents, \etc), allowing it to work in unconstrained \emph{open world} environments. OpenStereoNet is made of a convolutional Feature-Net for feature extraction, a Match-Net for depth prediction, and two recurrent Long Short-Term Memory (LSTM) blocks to encode and to exploit temporal dynamics in the video.  Importantly and in contrast to existing deep stereo matching methods, our network does not need any ground-truth disparity map as supervision, yet it naturally generalizes well to unseen datasets. As new videos are processed, the network is able to memorize, and to learn from, its past experiences.  Without needing ground-truth disparity maps, our network is able to tune its parameters after seeing more images from stereo videos, simply by minimizing image-domain warping errors. Also, to better leverage the sequential information in the stereo video, we apply the Long Short-Term Memory (LSTM) module to the bottleneck of feature extraction and feature matching part of our network. In the later part of this paper, we demonstrate that our method can be applied to vary open-world scenarios such as indoor/outdoor scenes, different weather/light conditions and different camera settings with superior performance. Also ablation study concerning the effect of the LSTM modules is conducted.  Another novelty of this work is that: we adopt convolutional-LSTM \cite{Shi2015} (cLSTM) as the recurrent feedback module, and use it directly on a continuous video sequence harnessing the temporal dynamics of the video. To our knowledge, while RNN-LSTM has been applied to other video processing tasks (such as sequence captions, or human action recognition), it has not been used for stereo matching for video sequences.

\section{Related work}
Stereo matching is a classic problem in computer vision, and has been researched for several decades. There have been significant number of papers published on this topic (The reader is referred to some survey papers \eg, \cite{Scharstein2002,Geiger-Survey-2017}).  Below we only cite a few most recent deep-learning based stereo methods that we consider most closely related to the method to be described.

\textbf{Supervised Deep Stereo Matching.}  In this category, a deep network (often based on CNN, or Convolutional Neural Networks) is often trained to benefit the task of stereo matching in one of the following aspects: i) to learn better image features and a tailored stereo matching metrics (\eg, \cite{Zbontar2016,Luo2016});  ii) to learn better regularization terms in a loss function \cite{Seki2017CVPR}; and iii) to predict dense disparity map in an end-to-end fashion (\eg, \cite{Mayer2016CVPR,KendallMDHKBB17}). The learned deep features replace handcrafted features, resulting in more distinctive features for matching. End-to-end deep stereo methods often formulate the task as either depth values regression, or multiple (discrete) class classification.  DispNetC \cite{Mayer2016CVPR} is a new development, which directly computes the correspondence field between stereo images by minimizing a regression loss.  Another example is the GC-Net \cite{KendallMDHKBB17}밃 which explicitly learns feature cost volume, and regularization function in a network structure.  Cascade residual learning (CRL) \cite{pang2017cascade} adopted a multi-stage cascade CNN architecture, following a coarse-to-fine or residual learning principle \cite{ResNet:CVPR-2016}.

\textbf{Unsupervised Deep Stereo Matching.}  Recently, there have been proposed deep net based single-image depth recovery methods which do not require ground-truth depth maps. Instead, they rely on minimizing photometric warping error to drive the network in an unsupervised way (see \eg,\cite{garg2016unsupervised,monodepth17,zhou2017unsupervised,Xie2016,SsSMnet2017}).
Zhou \etal \cite{zhou2017unsupervised} proposed an unsupervised method which is iteratively trained via warping error propagating matches. The authors adopted TV (total variation) constraint to select training data and discard uninformative patches. Inspired by recent advances in direct visual odometry (DVO), Wang \etal \cite{Chaoyang-CVPR-2018} argued that the depth CNN predictor can be learned without a pose CNN predictor. Luo \etal \cite{Single-View-Stereo-Matching:CVPR-2018} reformulated the problem of monocular depth estimation as two sub-problems, namely a view synthesis procedure followed by standard stereo matching. However, extending these monocular methods to stereo matching is non-trivial. When feeding the networks with stereo pairs, their performances are even not comparable with traditional stereo matching methods \cite{monodepth17}.

\textbf{Recurrent Neural Net and LSTM.} Our method is based on RNN (with cLSTM as the feedback module), and directly applied to sequence input of stereo video harnessing the temporal dynamic nature of a continuous video.  To the best of our knowledge, where RNN-LSTM has been applied to other video based tasks (such as a sequence captions, action recognition), it has not been directly used for  stereo video matching, especially to exploit the temporal smoothness feature for improving stereo matching performance.

\section{Network Architecture}
In this section, we describe our new ``open-world'' stereo video matching deep neural network (for ease of reference, we call it {\em OpenStereoNet}).  The input to the network is a live continuous stereo video sequence of left and right image frames of $I_L^t$,$I_R^t,$ for $t = 1,2,...$.  The output is the predicted depth-map (disparity map) at each time step $t$. We assume the input stereo images are already rectified.

Our network does not require ground-truth depth-maps as supervision. Instead, the stereo matching task is implemented by searching a better depth map which results in minimal photometric warping error between the stereo image pair. By continuously feeding in new stereo image frames, our network is able to automatically adapt itself to new inputs (could be new visual scenes never seen before) and produce accurate depth map estimations.  More technical details will be explained in the sequel of the paper.

\subsection{Overall network architecture}
\begin{figure}[!h]
\begin{center}
\includegraphics[width=0.8\linewidth,height=0.3\linewidth]{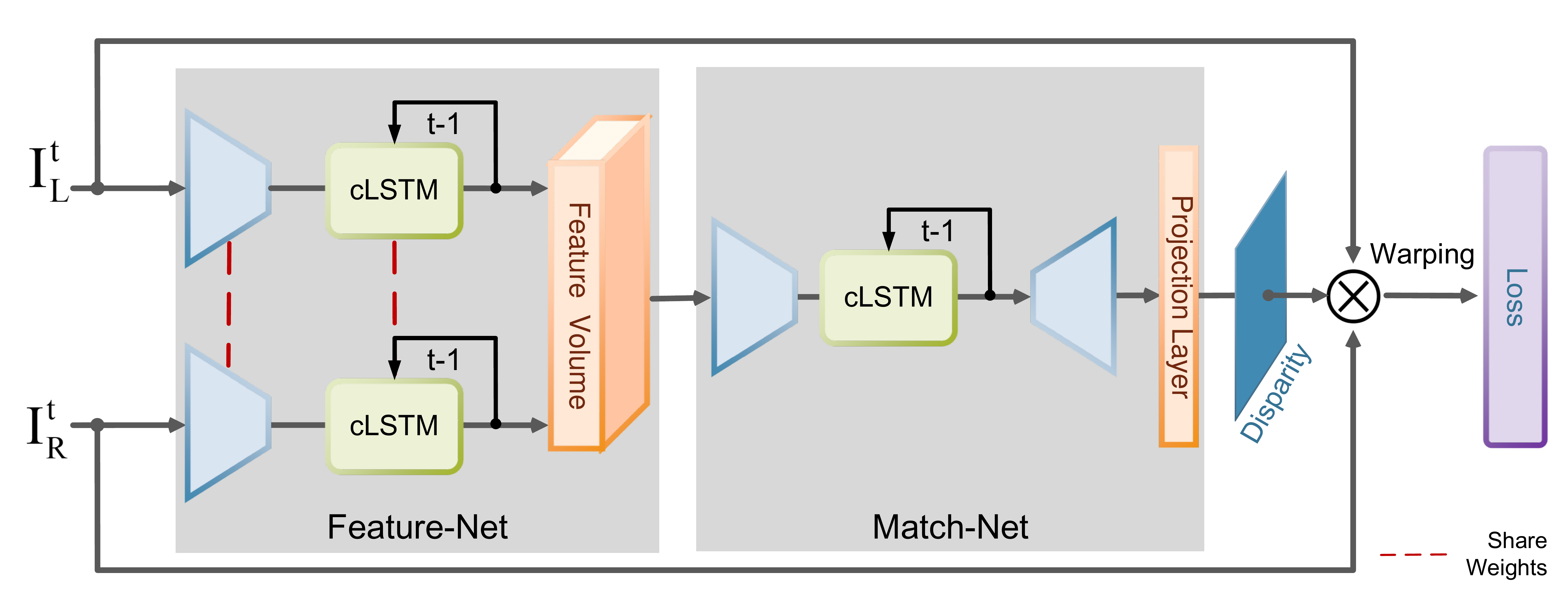}
\caption{\label{fig:structure}\small\textbf{Overall network structure of our OpenStereoNet.} It consists of a convolutional Feature-Net, an encoder-decoder type Match-Net and two recurrent (convolutional) LSTM units to learn temporal dynamics of the video input. Given a stereo pair $I_L^{t}, I_R^{t}$ at time $t$, the Feature-Net produces feature maps which are subsequently aggregated to form a feature-volume. The Match-Net first learns a representation of the feature volume then projects it to obtain disparity estimation. Our loss function is based on image warping error evaluated on raw image inputs by the current disparity map.}
\end{center}
\end{figure}

The overall structure of our OpenStereoNet is illustrated in Figure-\ref{fig:structure}밃 which consists of the following major parts (or sub-Nets): (1) Feature-Net, (2) Match-Net, (3) LSTM blocks, and (4) a loss function block.

\paragraph{Information Flow.} Starting from inputted left and right images at time $t$,  the information processing flow in our network is clear:  1) The Feature-Net acts as a convolutional feature extractor which extracts features from the left and right images individually. Note, Feature-Net for the left image and Feature-Net for the right image share the weights. 2) The obtained feature maps are concatenated (with certain interleave pattern) into a 4D feature-volume. 3) The Match-Net takes the 4D feature volume as input, and learns an encoder-decoder representation of the features. A projection layer (based on soft-argmin \cite{KendallMDHKBB17}) within the Match-Net is applied to produce the 2D disparity map prediction.  Finally, the loss function block employs the current estimated disparity map to warp the right image to the left view and compare the photometric warping loss as well as other regularization term, which is used to refine the network via {\em backprop}.

\subsection{Feature-Net}
Conventional stereo matching methods often directly compare the raw pixel values in the left image with that in the right image.  Recent advance in deep learning show that using learned convolutional features can be more robust for various vision tasks. For stereo matching, to learn a feature map that is more robust to photometric variations (such as occlusion, non-lambertian, lighting effects and perspective effects) will be highly desirable.

In this paper, we design a very simple convolutional feature-net with 18 convolutional layers (including RELU) using $3\times 3$ kernels and skip connections in between. The output feature has a dimensionality of 32.  We run feature extraction on both images in a symmetric weight-sharing manner.

\subsection{Feature-Volume construction}
\begin{figure}[!htp]
\begin{center}
\includegraphics[width=0.4\linewidth]{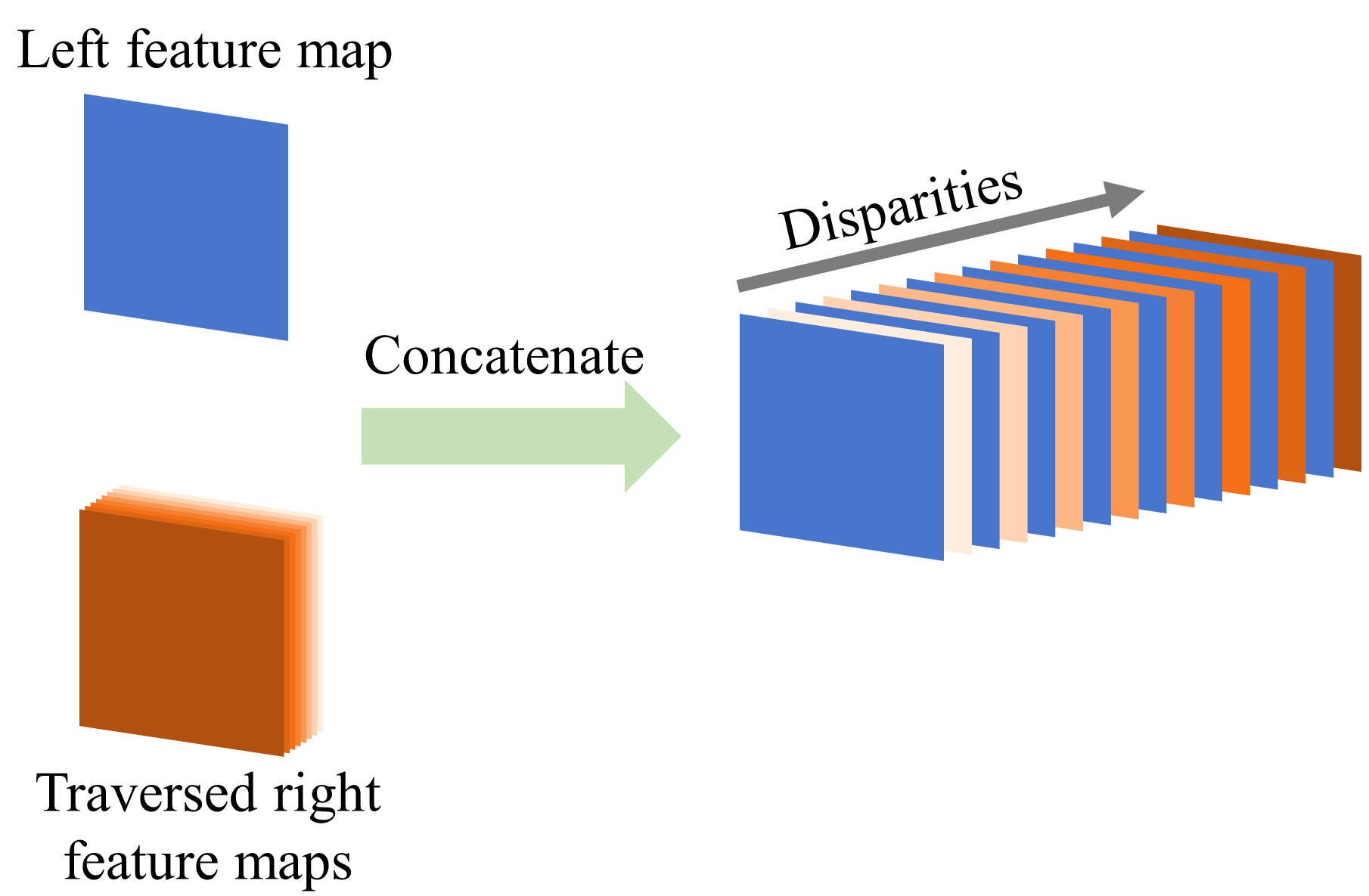}
\caption{\label{fig:vol}\small \textbf{Feature Volume Construction.}  We collect the two feature maps computed by the Feature-Net, and assemble them together to a Feature-Volume in the way as illustrated here: the blue rectangle represents a feature map from the left image, the stacked orange rectangles represent traversed the right feature at different disparities in the range $[0,D]$.  Note that the left feature map is duplicated $D+1$ times to match the traversed right feature maps.}
\end{center}
\end{figure}

We use the learned features to construct a feature volume. Instead of constructing a cost volume by concatenating all costs with their corresponding disparities, we concatenate the learned features from the left and right images. Specifically, we concatenate each learned feature with their corresponding feature from the opposite stereo image across each disparity level in a preset disparity range $D$ as illustrated in Fig.~\ref{fig:vol}. All the features are packed to form a 4D feature volume with dimensionality $H\times W\times(D+1)\times 2F$ for the left-to-right and right-to-left feature volume correspondingly, where $H,W,D,F$ represent the height, width, disparity range, and feature dimensionality respectively.

\subsection{Match-Net}
Taking the assembled Feature-Volume as input, our Match-Net is constituted of an encoder-decoder as the front-end, followed by a single last layer which projects the output of the encoder-decoder to a 2D disparity-map.

\subsubsection{Encoder-Decoder front-end.}
The denoising Encoder-Decoder is an hour-glass-shaped deep-net. Between the encoder and decoder there is a bottleneck, as shown in Figure-\ref{fig:structure}.  Since the input feature-volume is of 4 dimensions, H ({height}) $\times$ W ({width}) $\times$ (D+1)({disparity range}) $\times$ 2F({feature ~dim.}), we use 3D-convolutional kernels and the underlying CNNs in the Encoder-Decoder are in fact 3D-CNNs.

\subsubsection{Projection layer.} The output of the preceding encoder-decoder is still a 4D feature-volume. The last layer of our Match-Net first projects the 4D volume to a 3D {\em cost-volume}--\ie an operation commonly used in conventional stereo matching methods, then applies the soft-argmin operation (\cf \cite{KendallMDHKBB17}) to predict a disparity $\delta =\sum_{d=0}^D [d \times \sigma(-c_d)]$, where $c_d$ is the matching cost at disparity $d$ and $\sigma(\cdot)$ denotes the softmax operator.

\subsection{Convolutional-LSTM}
Since our goal is to develop a deep-net focusing on stereo video processing (as opposed to individual images), in order to capture the inherent temporal dynamics (\eg, temporal smoothness) existed in a video, we leverage the internal representations obtained by our two sub-networks (\ie, Feature-Net and Match-Net), and model these internal representation's dynamic transitions as an {\em implicit} model for the video sequence. Specifically, given a continuous video, we consider the image content (as well as the disparity) in each frame changes smoothly to the next frame. To capture such dynamic changes, we adopt the structure of LSTM-based Recurrent Neural Networks (RNN). The LSTMs act as memory of the net, by which the network memorizes its past experiences. This gives our network the ability to learn the stereo video sequence's temporal dynamics encoded in the inner states of the LSTMs.
\begin{wrapfigure}{L}{0.4\textwidth}
\centering
\includegraphics[width=0.33\textwidth]{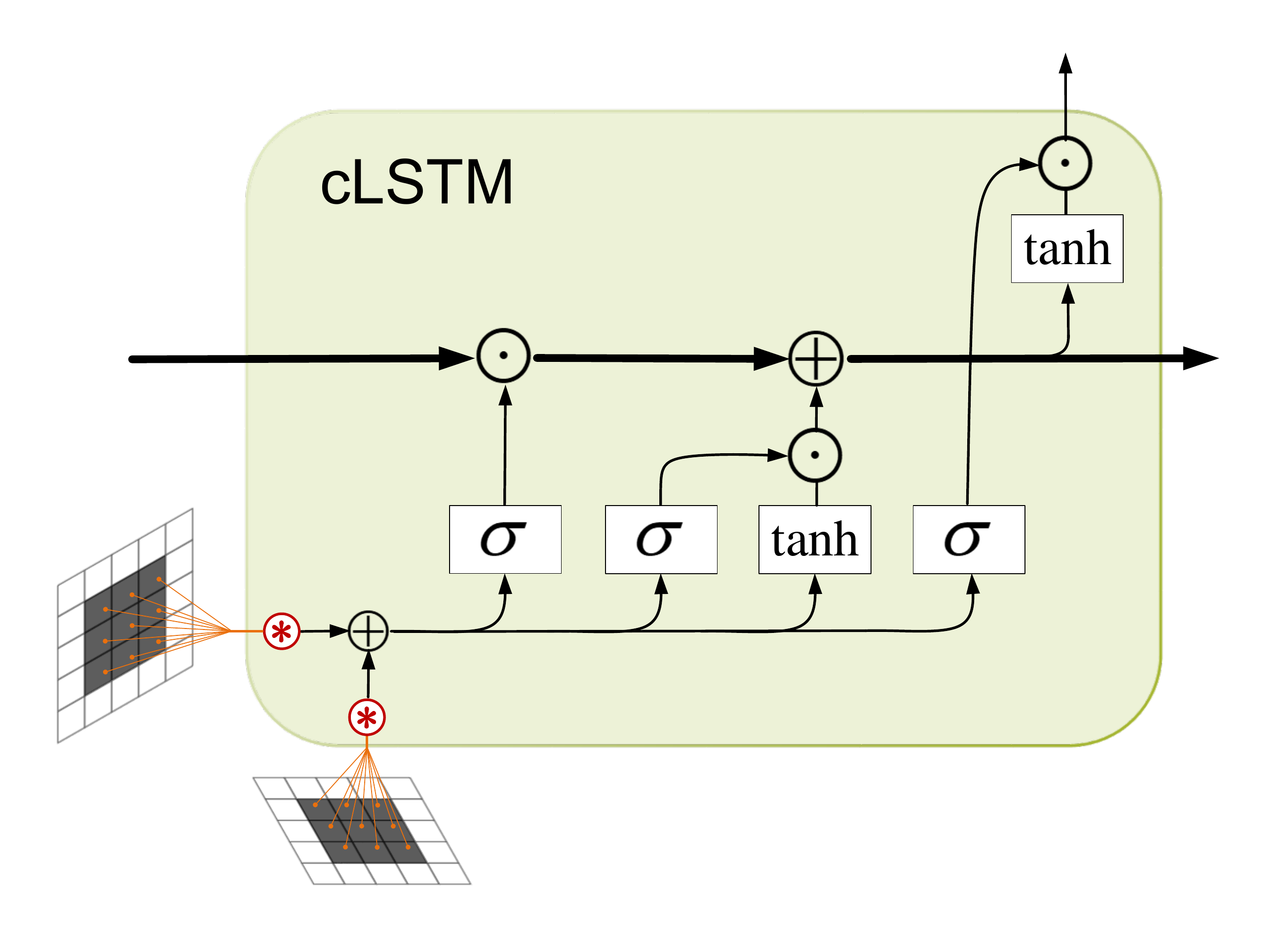}
\caption{\label{fig:cLSTM}\small \textbf{A convolutional-LSTM.}}
\end{wrapfigure}
As shown in Fig.-\ref{fig:cLSTM}, the output of our Feature-Net, and the encoder-decoder in our Match-Net, are each passed to an LSTM unit. Briefly, LSTM units are a particular type of hidden unit that improve the training of RNNs \cite{Bengio1994}.  An LSTM unit contains a cell,  which can be thought of as a memory state. Access to the cell is controlled through an input gate and a forget gate.  The final output of the LSTM unit is a function of the cell state and an output gate (\cf \cite{Schmidhuber:LSTM-1997}).

We realize that for stereo matching it is desirable to use spatially-invariant operator such as convolutional kernels, in the same spirit of the CNN.  In light of this, we propose to use the convolutional LSTM architecture (or cLSTM in short, \cf \cite{Shi2015}) as the recurrent unit for our stereo video matching task. This way, the relative spatial layout information of the feature representation--which is essential for the task of depth map computation--is preserved.  In our experiments we used very small kernels for the cLSTM (\eg $3\times 3$, or $5\times 5$). This leads to compact LSTM units, and greatly simplifies the computation cost and GPU-RAM consumption.  The encoder--LSTM--decoder architecture is also similar to the Encoder-Recurrent-Decoder architecture proposed in \cite{Fragkiadaki:Recurrent-Model-ICCV-2015}.  Our entire network works end-to-end to combine feature representation learning with the learning of temporal dynamics via the two LSTM blocks.

\section{Self-Adapting Learning and Loss Function}
\subsection{Self-Adapting Training and Testing}
Recall that the ultimate goal of this work is to develop a deep network that can automatically adapt itself to new (previously unseen) stereo video inputs.  In this sense, the network is not fixed static, but is able to dynamically evolve in time. This is achieved by two mechanisms:
\begin{itemize}
\item The network has memory units (i.e. the two LSTM blocks), which enable the network to adjust its current behavior (partly) based its past experiences;
\item We always run an online back-propagation (\emph{backprop}) updating procedure after any feed-forward process.
\end{itemize}
The latter actually eliminates the separation between a network's training stage and testing stage.  In other words, our OpenStereoNet is constantly performing both operations all the time. This gives the network self-adaption ability, allows it to self-adapt by continuously fine-tuning its parameters based on new stereo image inputs (possibly seen in a new environment).  Thus, our OpenStereoNet can ``automatically'' generalize to unseen images.

Since we do not require ground-truth depth-maps as supervision, input stereo pairs themselves serve as self-supervision signals, and the network is able to update automatically, by self-adapting learning.

\subsection{Overall loss function}
The overall loss function for our OpenStereoNet is a weighted summation of a data term and a regularization term, as in $Loss = L_{\rm data} + \mu L_{\rm reg}.$

\textbf{Data term: Image warping error.}
We directly measure the warping error evaluated on the input stereo images, based on the estimated disparity map. Specifically, given the left image $I^t_L$ and the disparity map for the right image $d^t_R = g(I^t_R, I^t_L,h^{t-1}_R,h^{t-1}_L)$, the right image $I^t_R$ can be reconstructed by warping the left image with $d^t_R$, $I_R^{t'}(u,v) = I^t_L(u+d,v),$ where $I_R^{t'}$ is the warped right image. We use the discrepancy between $I_R^{t'}$ and the observed right image $I^t_R$ as the supervision signal. Our data loss is derived as:
${L}_{\rm data}= \sum(\lambda_1(1-\mathcal{S}(I_L, I_L^{'}))/2 + \lambda_2(\left| I_L- I_L^{'}\right|+\left| \nabla I_L - \nabla I_L^{'}\right|))/N$. The data term consists of $S$, which is the structural similarity SSIM as defined in \cite{SSIM2004}, pixel value difference and image gradient difference.  The trade-off parameters were chosen empirically in our experiments at $\mu=0.05, \lambda_1 =0.8, \lambda_2=0.1.$

\textbf{Regularization term: Priors on depth-map.}
We enforce a common prior that depth-maps are piecewise smooth or piecewise linear.  This is implemented by penalizing the second-order derivative of the estimated disparity map.  To exploit correlations between depth map values and pixel colors, we weight this term by image color gradient, \ie : ${L}_{\rm reg}= \sum(e^{-\left|\nabla^2_u I_L\right|}\left|\nabla^2_u d_L\right|+ e^{-\left| \nabla^2_v I_L\right|}\left| \nabla^2_v d_L\right|)/N,$ where $\nabla$ is gradient operator.

\section{Experiments}
We implement our OpenStereoNet in TensorFlow. Since the network runs in an online fashion (with batch-size one), \ie, there is no clear distinction between training and testing, we start from randomly initialized weights for both the Feature-Net and the Match-Net, and allow the network to evolve as new stereo images being fed in. All images have been rescaled to $256\times512$ for easy comparison. Typical processing time of our net is about 0.8--1.6 seconds per frame tested on a regular PC of 2017 equipped with a GTX 1080Ti GPU. We use the RMSProp optimizer with a constant learning rate of $0.001$.
We have evaluated our network on several standard benchmark datasets for stereo matching, including KITTI\cite{Geiger2012CVPR}, Middlebury\cite{Scharstein07,HirschmullerS07}, Synthia \cite{RosCVPR16}, and Frieburg SceneFlow \cite{Mayer2016CVPR} (e.g. FlyingThings3D). These experiments are reported below.
\subsection{KITTI visual odometry (VO) stereo sequences}
In this set of experiments on KITTI dataset \cite{Geiger2012CVPR}, we simply feed a KITTI VO stereo video sequence to our network, and start to produce a depth map prediction, as well as update the network weights frame by frame by backproping the error signal of the loss function.  In all our experiments we observe that: soon after about a few hundreds of input frames have been processed (usually about 10 seconds video at 30fps) the network already starts to produce sensible depth maps, and the loss function appears to converge.
We call this process of training on the first a few hundred frames the {\em network prime} process, and we believe its purpose is to teach the network to learn useful convolutional features for typical visual scenes.  Once the network has been ``primed'', it can be applied to new previously unseen stereo videos.

\begin{figure}[!h]
\begin{center}
\subfigure{
\includegraphics[width=0.4\linewidth,height=0.2\linewidth]{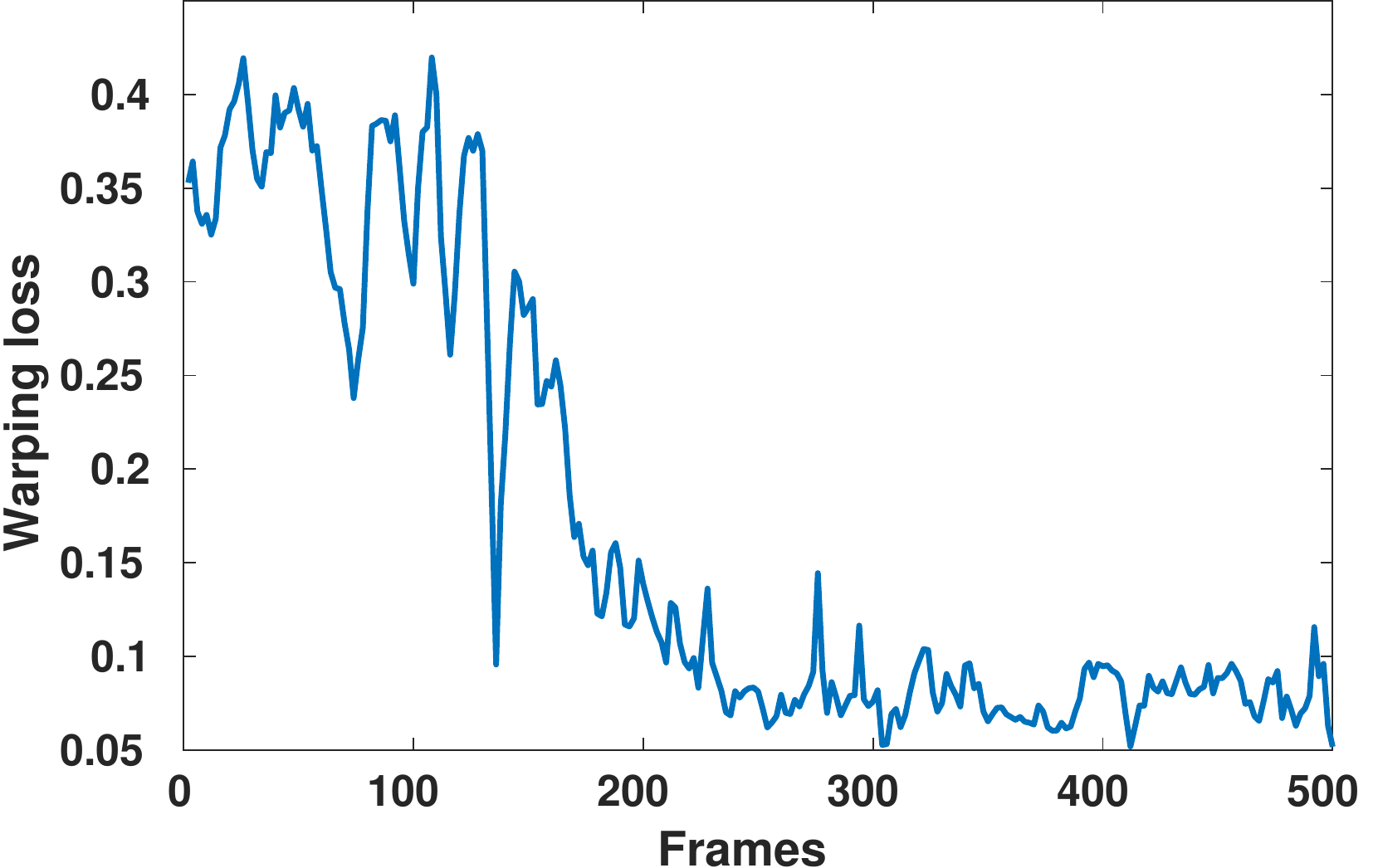} }
\subfigure{
\includegraphics[width=0.4\linewidth,height=0.2\linewidth]{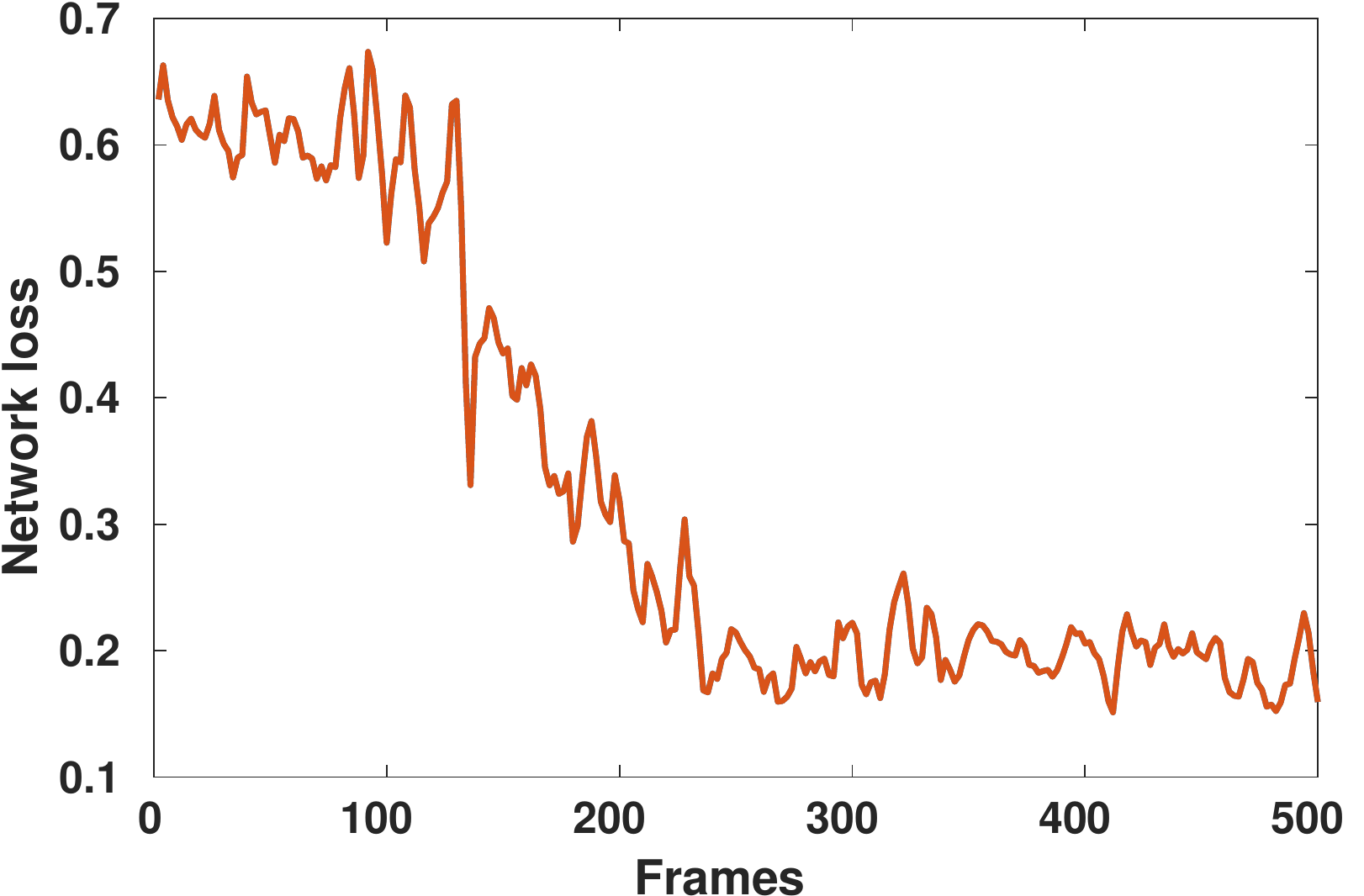} }
\caption{\label{fig:inital}\small \textbf{Typical network convergence curves from random initialization.}}
\end{center}
\end{figure}

Figure-\ref{fig:inital} shows typical converge curves for a network during the prime stage.
After the prime stage, we randomly select 5 KITTI VO sequences, test our network on them, and compare its performance with three state-of-the-art stereo methods, including the DispNet \cite{Mayer2016CVPR}, MC-CNN \cite{Zbontar2016}, and SPS-ST \cite{Yamaguchi14}. The first two are deep stereo matching methods, and the last one a traditional (non-deep) stereo method. Quantitative comparison of their performances are reported in Table-\ref{tab:kitti_vo}, from which one can clearly see that our OpenStereoNet achieves the best performance throughout all the metrics evaluated.  For deep MC-CNN we use a model which was firstly trained on Middlebury dataset for the sake of fair comparison.  For SPS-ST, its meta-parameters was also tuned on KITTI dataset.
\begin{figure}[!ht]
\begin{center}
\subfigure[Left frame]{
\includegraphics[width=0.3\linewidth]{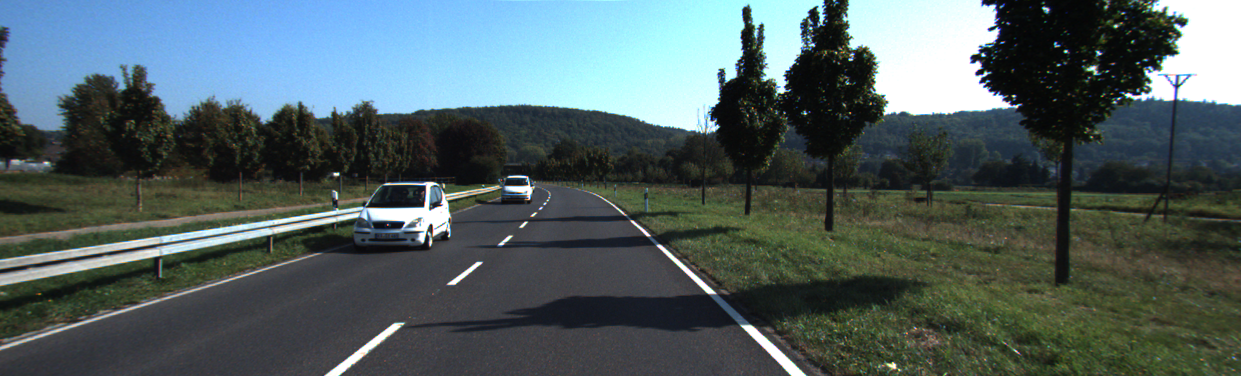} }
\subfigure[Sparse LIDAR]{
\includegraphics[width=0.3\linewidth]{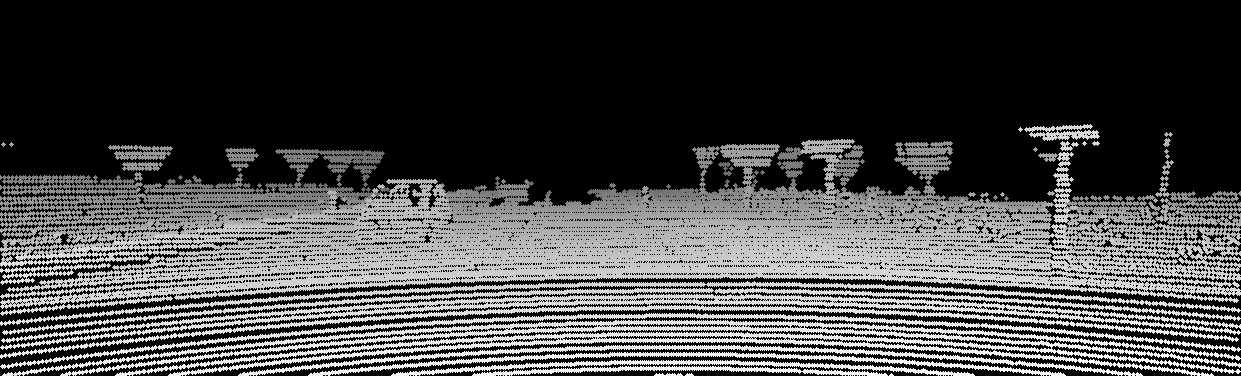} }
\subfigure[Ours ($3.44\%$)]{
\includegraphics[width=0.3\linewidth]{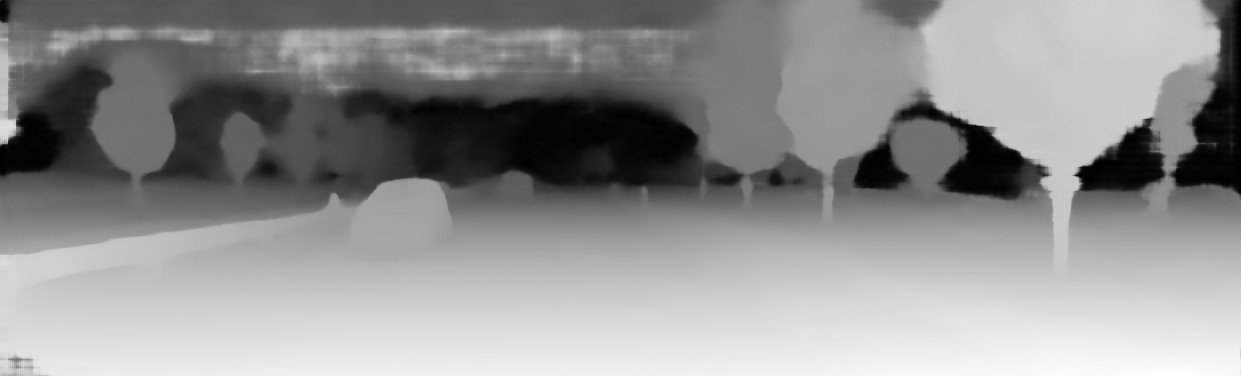} }
\subfigure[SPS-st ($5.61\%$)]{
\includegraphics[width=0.3\linewidth]{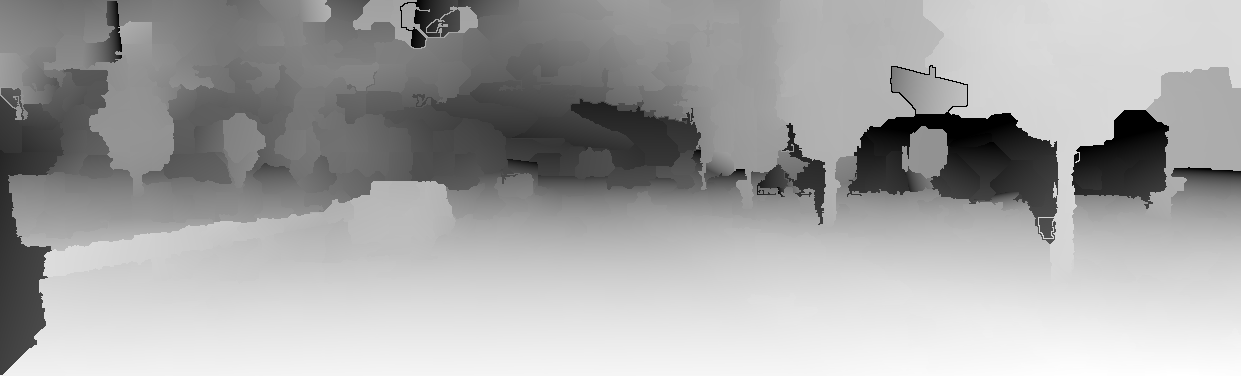} }
\subfigure[MC-CNN ($4.75\%$)]{
\includegraphics[width=0.3\linewidth]{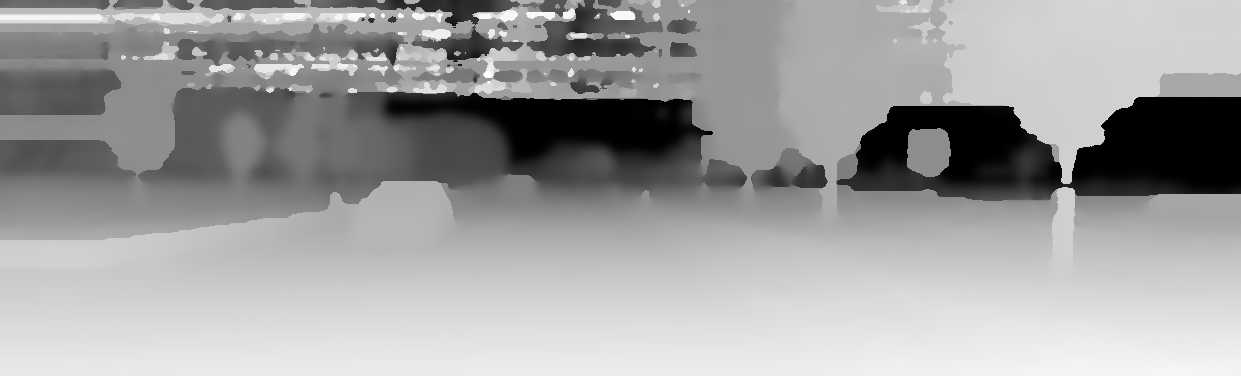} }
\subfigure[DispNet ($25.98\%$)]{
\includegraphics[width=0.3\linewidth]{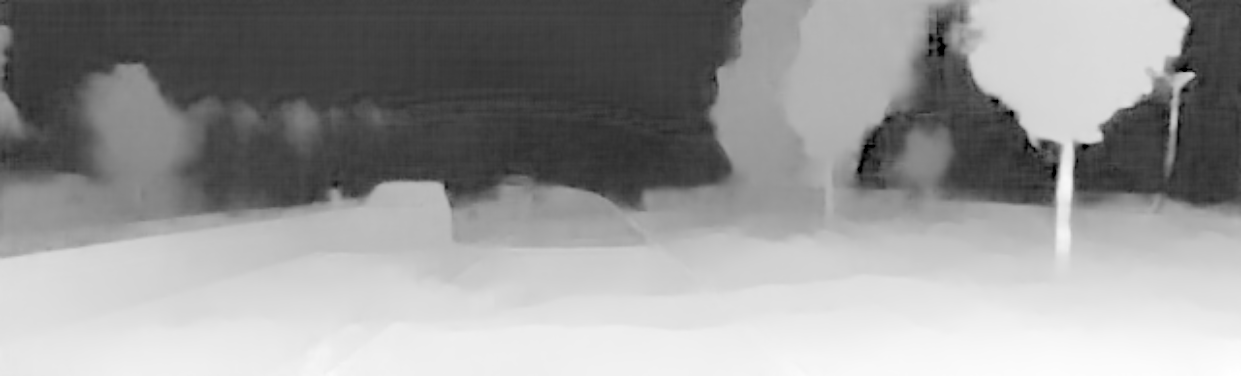} }
\caption{\label{fig:result_vo}\small \textbf{Our qualitative results on KITTI VO dataset:} Results are reported on the $D1\_{all}$ error metric. Disparities are transformed to log space for better visualization.}
\end{center}
\end{figure}
Figure--\ref{fig:result_vo} gives some sample visual results for comparison.  Note that our method obtains sharp and clean depth discontinuity for cars and trees,  better than the other methods. Dispnet, on the other hand, is affected by shadows on road, with many artifacts on the road surface. A quantitative comparison is provided in Table-\ref{tab:kitti_vo}. Our method outperforms all baseline methods with a large margin.

\begin{table}[!ht]
\small
\centering
\tabcolsep=0.04cm
\begin{tabular}{|c|c|c|c|c|c|c|}
\hline
Methods               & Abs Rel & Sq Rel  & RMSE & RMSE log모& D1\_all모& $\delta<1.25$  \\ \hline
Dispnet \cite{Mayer2016CVPR}   & 0.122  & 1.938 & 8.844 & 0.189 & 32.045 &0.877  \\
MC-CNN \cite{Zbontar2016}   & 0.069  & 1.229 & 6.002 & 0.264 & 8.018 &0.932 \\
SPS-st \cite{Yamaguchi14}   & 0.060  & 1.341 & 5.521 & 0.159 & 4.970 &0.957  \\
Ours  & {\bf 0.053}  & {\bf0.540} & {\bf 4.451} & {\bf 0.137} &  {\bf4.403} &  {\bf 0.959} \\ \hline
\end{tabular}
\caption{\label{tab:kitti_vo}\small {Quantitative results on KITTI VO dataset.}}
\end{table}

\subsection{Synthia Dataset}
\begin{figure}[!ht]
\begin{center}
\includegraphics[width=0.3\linewidth]{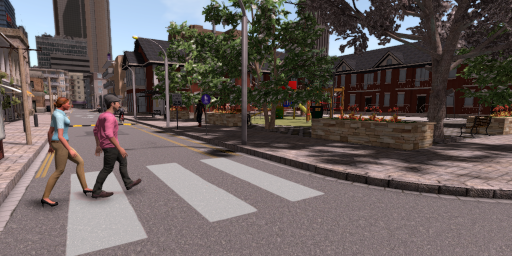}
\includegraphics[width=0.3\linewidth]{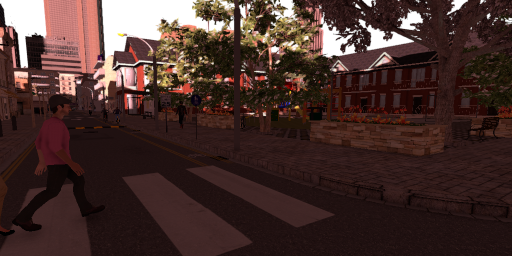}
\includegraphics[width=0.3\linewidth]{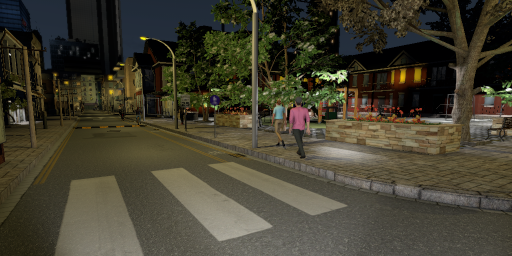} \\
\includegraphics[width=0.3\linewidth]{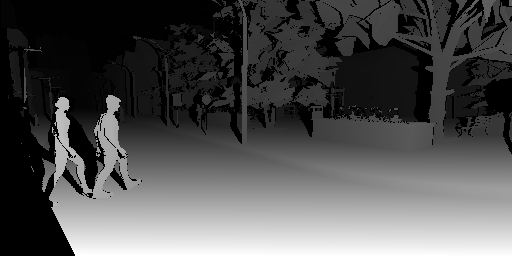}
\includegraphics[width=0.3\linewidth]{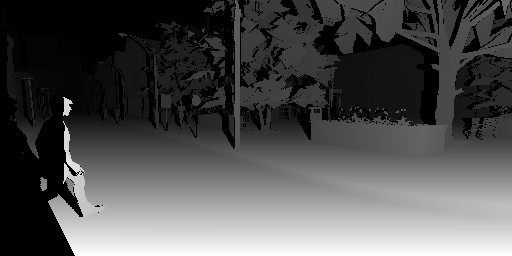}
\includegraphics[width=0.3\linewidth]{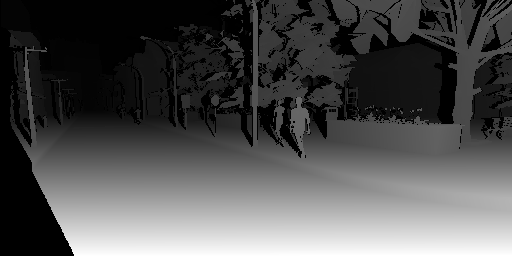} \\
\includegraphics[width=0.3\linewidth]{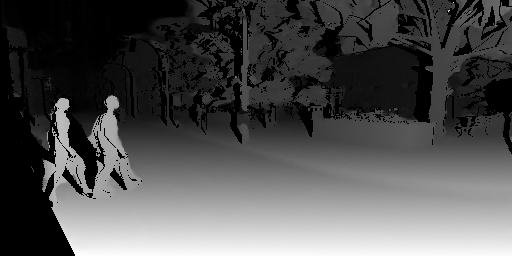}
\includegraphics[width=0.3\linewidth]{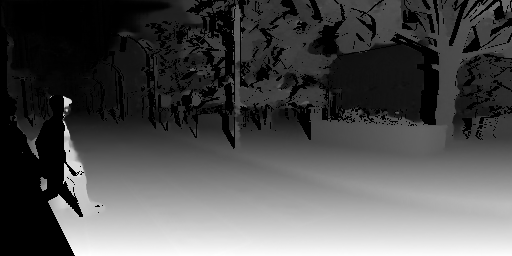}
\includegraphics[width=0.3\linewidth]{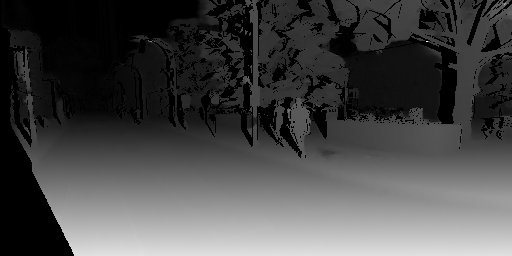}
\caption{\label{fig:result_sy}\small \textbf{Our qualitative results on Synthia:} Top to bottom: input left image, ground truth disparity, our result.  The first column is taken from the \emph{Spring} subset, the middle column is from \emph{Dawn}, and the last column is from \emph{Night}.  Our method performs uniformly on different sequences.}
\end{center}
\end{figure}

The Synthia dataset \cite{RosCVPR16} contains 7 sequences with different scenarios under different seasons and lighting conditions. Our primary aim for experimenting on Synthia is to analyze our network's generalization (self-adaption) ability.  We create a long video sequence by combining together three Synthia sequences of the same scene but under different seasons and lighting conditions.  For example, Fig.-\ref{fig:result_sy} shows some sample frames of \emph{Spring, Dawn and Night}. We simply run our network model on this video, and display the disparity error as a function of frames.  We run our network on this long sequence. For each condition, we report our quantitative and qualitative results based on the first 250 frames of that sequence.

\begin{figure}[!ht]
\begin{center}
\includegraphics[width=0.9\linewidth]{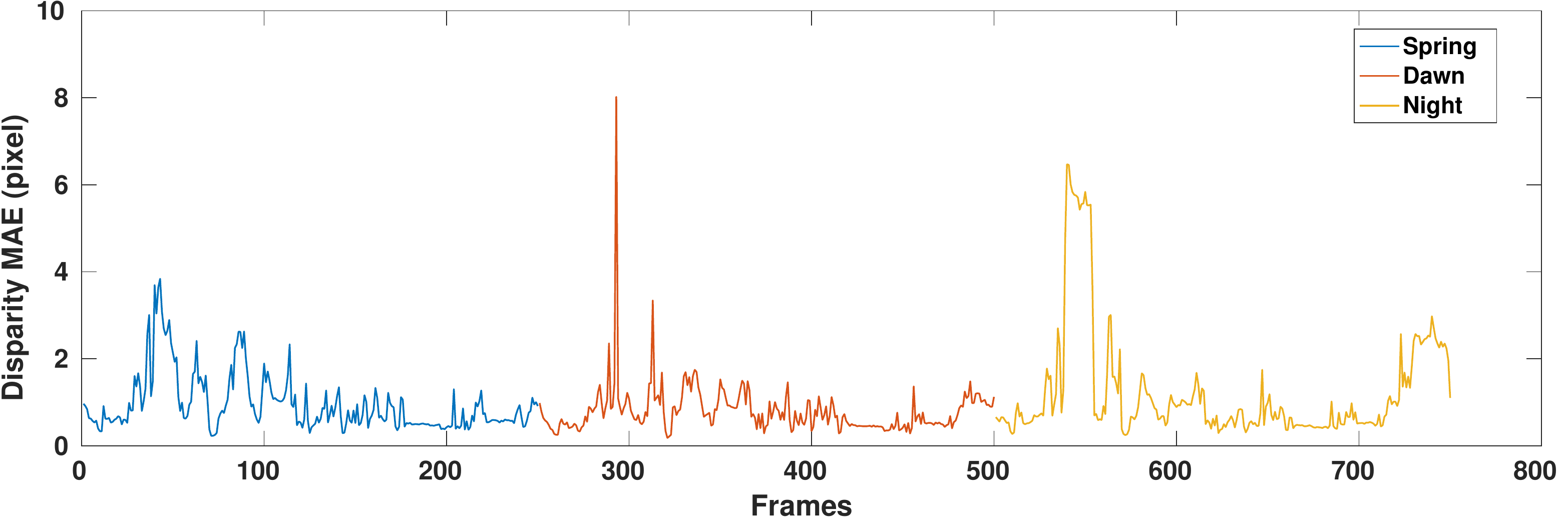}
\caption{\label{fig:synthia_mae}\small We run our method on a continuous video sequence consisting of the same scene under three different season/lighting conditions (spring, dawn, night).  The curve shows the final disparity error as a function of the stereo frame.  From this curve it is clear that our network is able to adapt to new scenarios automatically.}
\end{center}
\end{figure}

As shown in Fig.~\ref{fig:result_sy}, our network recovers consistently high quality disparity maps regardless the lighting conditions. This claim is further proved by the qualitative results in Fig.-\ref{fig:synthia_mae}. In term of disparity accuracy, our method achieves a Mean Absolute Error (MAE) of 0.958 pixels on the \emph{Spring} scene while the \emph{Dawn} sequence has reached an MAE of 0.7991 pixels and 1.2415 pixels for the \emph{Night} sequence.

\subsection{Ablation Studies: Effects of the LSTMs and Backprop}
There are two mechanisms that contribute to the self-adaptive ability of our OpenStereoNet, \ie, the cLSTMs recurrent blocks and the backprop refinement process.  To understand their respective effects on the final performance of our network, we conduct ablation studies by isolating their operations.  To be precise, we have tested the following four types of variants of our full networks:  (type-1) remove LSTMs and also disable the backprop process (\ie, the {\em baseline} network); (type-2) remove LSTMs, but keep backprop on; (type-3) with LSTMs on, without backprop, and (type-4) with both LSTMs on and backprop on (\ie, our full network).

\begin{figure}[!ht]
\begin{center}
\includegraphics[width=0.45\linewidth]{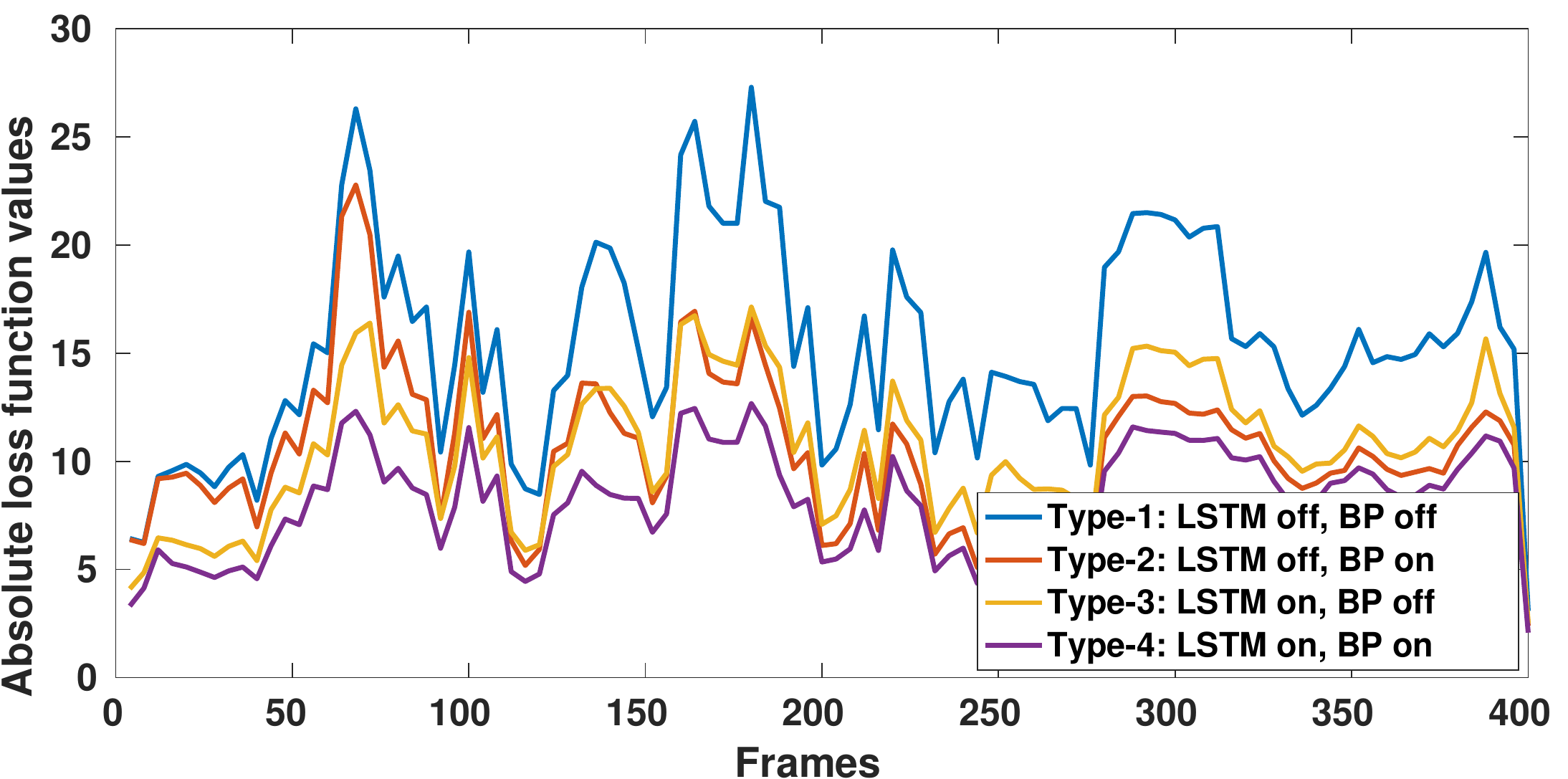} ~~~ \includegraphics[width=0.45\linewidth]{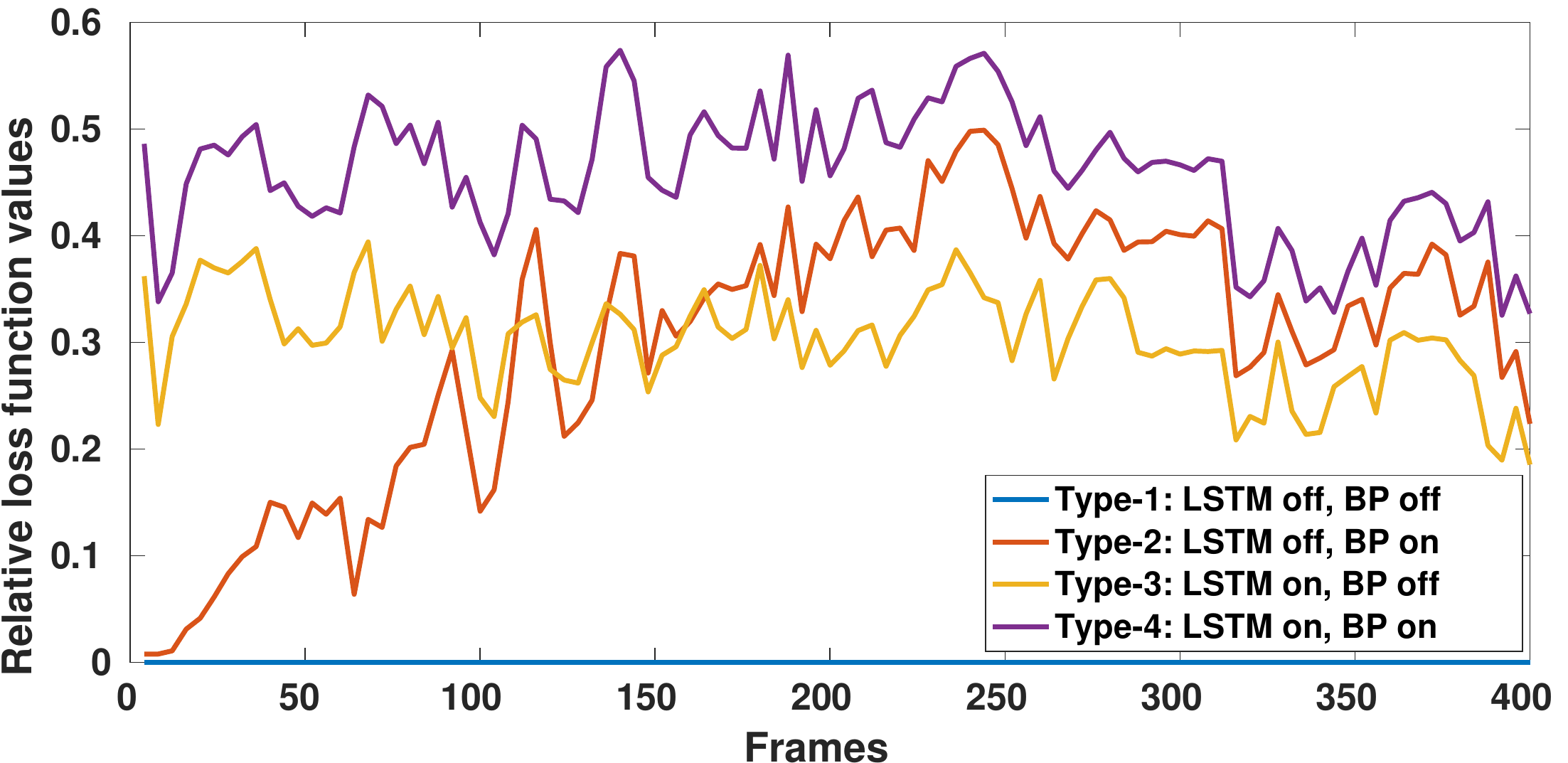}
\caption{\label{fig:lstmVSconv}{\small We run our method on a continuous video sequence. The left figure shows the comparison of absolute losses  for the 4 types (note: the lower, the better),  while the right figure gives comparison of the relative loss against the the (type-1) baseline network (note: the higher, the better). The y-axis indicates the final disparity errors, and x-axis the input frame-Ids.}}
\end{center}
\end{figure}

Results by these four types of networks are given in the following curves in Figure-\ref{fig:lstmVSconv}.  One can clearly see the positive effects of the LSTM units and the backprop. In particular, adding LSTMs has reduced the loss function of the baseline network significantly.
\begin{table}[!ht]
\small
\centering
\tabcolsep=0.03cm
\begin{tabular}{|c|c|c|c|c|c|c|}
\hline
Methods               & Abs Rel & Sq Rel  & RMSE & RMSE log모& D1\_all모& $\delta<1.25$  \\ \hline
Type-3 net (without LSTMs)  & 0.066  & 1.580 & 5.332 & 0.167 & 5.089 &0.957 \\
Type-4 net ( with LSTMs  )  & {\bf 0.053}  & {\bf0.540} & {\bf 4.451} & {\bf 0.137} &  {\bf4.403} &  {\bf 0.959} \\ \hline
\end{tabular}
\caption{\label{tab:kitti_lstmvsconv}\small {Ablation study on LSTM module on KITTI.}}
\end{table}
In another ablation test, we run the above type-3 network on the previous selected KITTI VO sequence, and list their accuracy in Table- \ref{tab:kitti_lstmvsconv}. From this, one can see that by applying the LSTM module, we have achieved better performance across all error metrics.

\subsection{Middlebury Stereo Dataset}
The stereo pairs in the Middlebury stereo dataset \cite{Scharstein07,HirschmullerS07} are indoor scenes with multiple handcrafted layouts. The ground truth disparities are captured by structured light with higher density and precision than KITTI dataset. We report our results on selected stereo pairs from Middlebury 2005 \cite{Scharstein07} and 2006 \cite{HirschmullerS07} and compare with other baseline methods. In order to evaluate our method on these images, we augmented each stereo pair to a stereo video sequence by simply repeating the stereo pair.

\begin{figure}[!ht]
\begin{center}
\subfigure[Aloe]{
\includegraphics[width=0.14\linewidth]{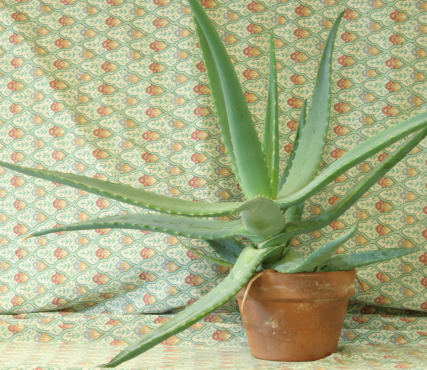} }
\subfigure[GT $0.00\%$]{
\includegraphics[width=0.14\linewidth]{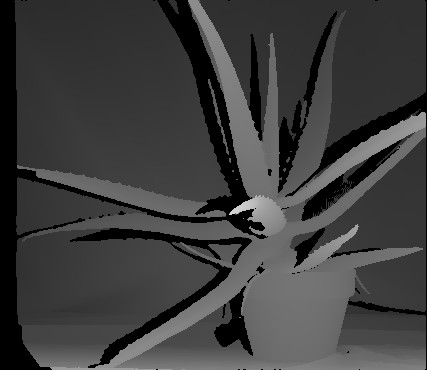} }
\subfigure[Ours $4.34\%$]{
\includegraphics[width=0.14\linewidth]{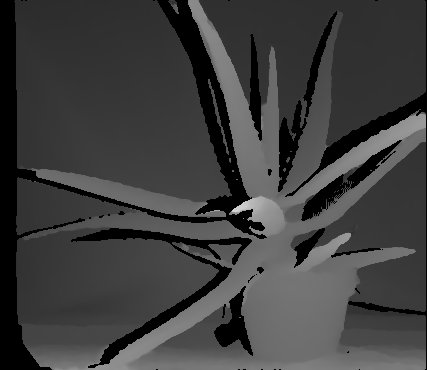} }
\subfigure[SPS-st $8.57\%$]{
\includegraphics[width=0.14\linewidth]{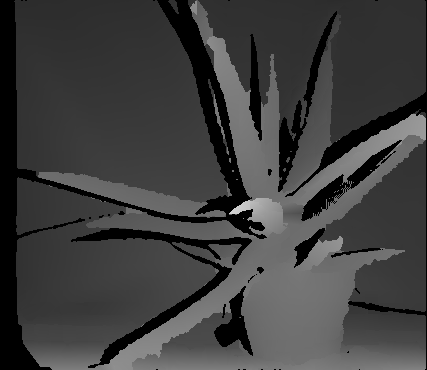} }
\subfigure[MC-CNN $16.72\%$]{
\includegraphics[width=0.14\linewidth]{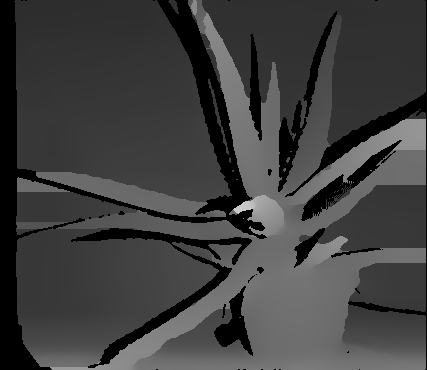} }
\subfigure[DispNet $35.77\%$]{
\includegraphics[width=0.14\linewidth]{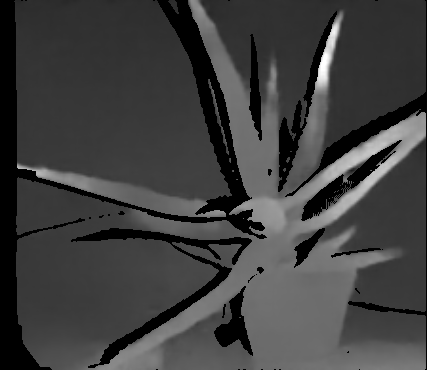} }
\subfigure[Dolls]{
\includegraphics[width=0.14\linewidth]{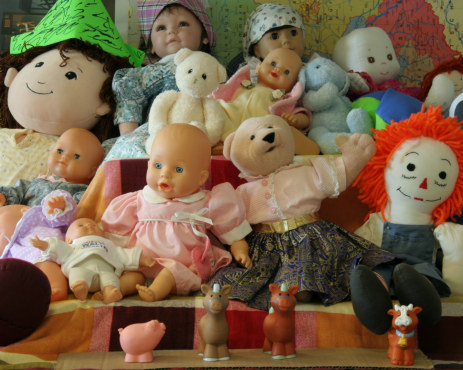} }
\subfigure[GT $0.00\%$]{
\includegraphics[width=0.14\linewidth]{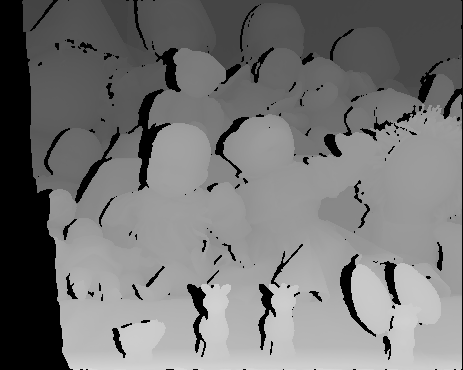} }
\subfigure[Ours $6.88\%$]{
\includegraphics[width=0.14\linewidth]{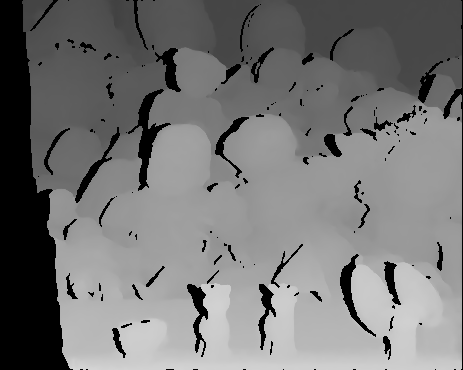} }
\subfigure[SPS-st $15.54\%$]{
\includegraphics[width=0.14\linewidth]{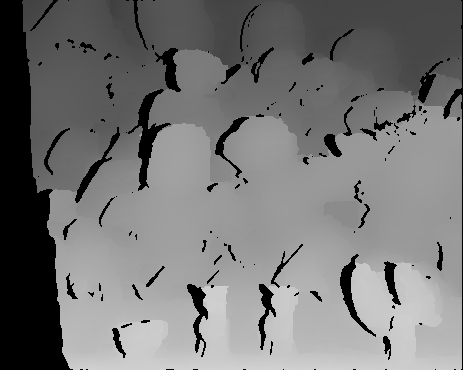} }
\subfigure[MC-CNN $23.78\%$]{
\includegraphics[width=0.14\linewidth]{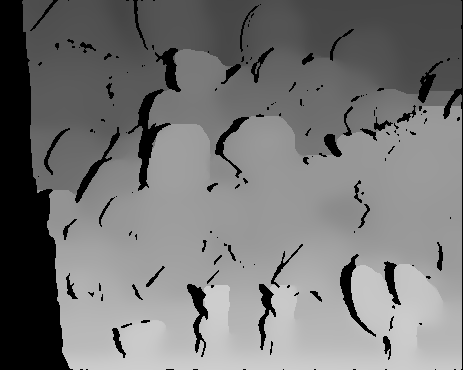} }
\subfigure[DispNet $44.52\%$]{
\includegraphics[width=0.14\linewidth]{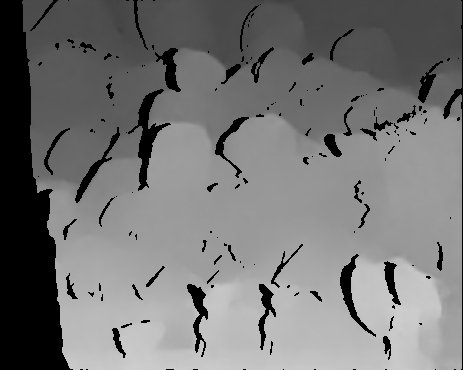} }
\caption{\label{fig:result_mb}\small \textbf{Our results on Middlebury:} Left to right: Left image, ground-truth disparity, our result, SPS-st result, MC-CNN result and DispNet result. We report the {\em bad-pixel-ratio} at 1-pixel threshold.}
\end{center}
\end{figure}

We use {\em bad-pixel-ratio} as our error metrics used in this experiment, and all results are reported with 1-pixel thresholding. As shown in Fig.~\ref{fig:result_mb}, our method achieves superior performance than all baseline methods. Other deep learning based methods have even worse performance than the conventional method SPS-st when there is no fine tuning.

\subsection{Other open world stereo sequences}
To further demonstrate the generalization ability of our OpenStereoNet, we test it on a number of other freely downloaded stereo video datasets from the Internet.  Note that our network had never seen these test data before.   Below we give some sample results,  obtained by our method and by the DispNet, on the Freiburg Sceneflow Dataset \cite{Mayer2016CVPR} and on RDS-Random Dot Stereo.

\begin{figure}[!ht]
\begin{center}
\subfigure{
\includegraphics[width=0.22\linewidth]{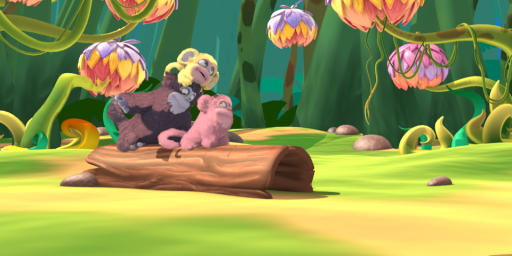} }
\subfigure{
\includegraphics[width=0.22\linewidth]{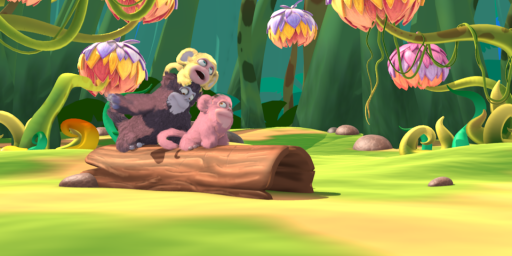} }
\subfigure{
\includegraphics[width=0.22\linewidth]{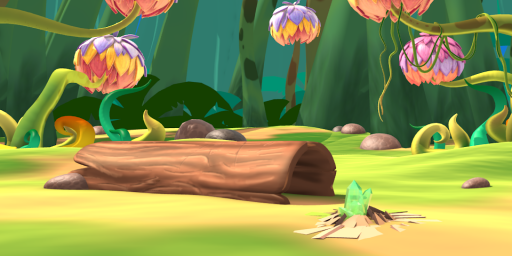} }
\subfigure{
\includegraphics[width=0.22\linewidth]{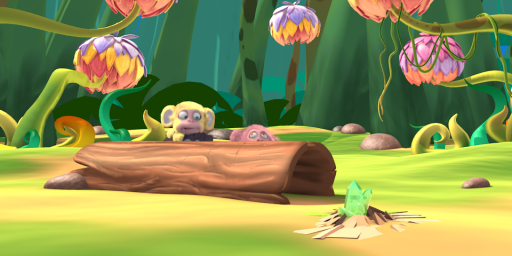} }
\subfigure{
\includegraphics[width=0.22\linewidth]{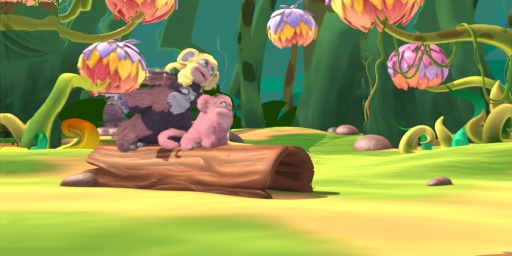} }
\subfigure{
\includegraphics[width=0.22\linewidth]{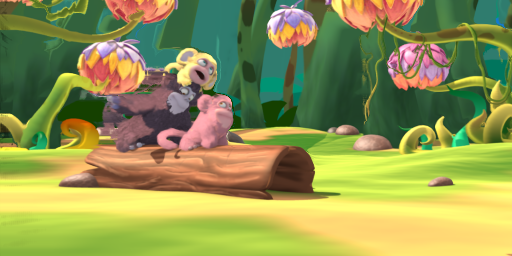} }
\subfigure{
\includegraphics[width=0.22\linewidth]{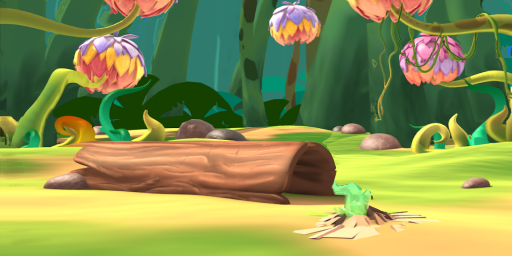} }
\subfigure{
\includegraphics[width=0.22\linewidth]{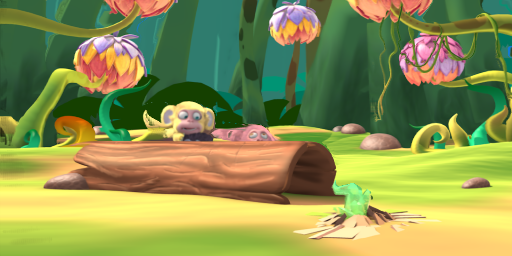} }
\subfigure{
\includegraphics[width=0.22\linewidth]{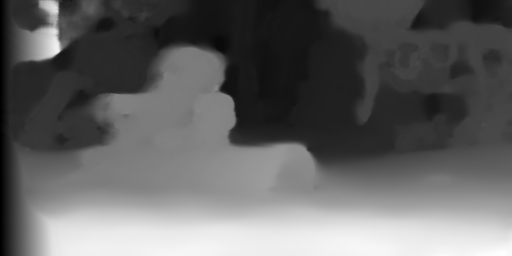} }
\subfigure{
\includegraphics[width=0.22\linewidth]{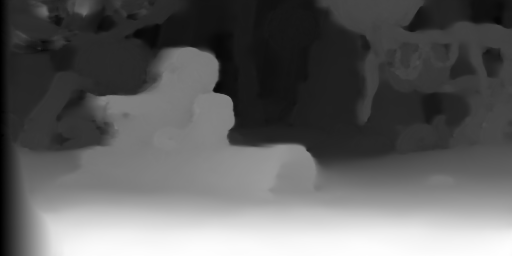} }
\subfigure{
\includegraphics[width=0.22\linewidth]{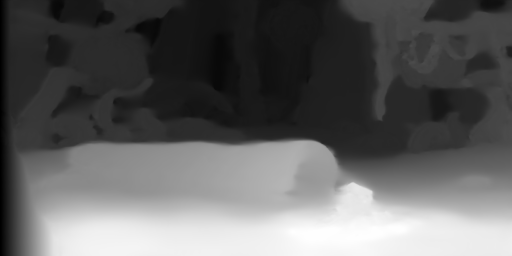} }
\subfigure{
\includegraphics[width=0.22\linewidth]{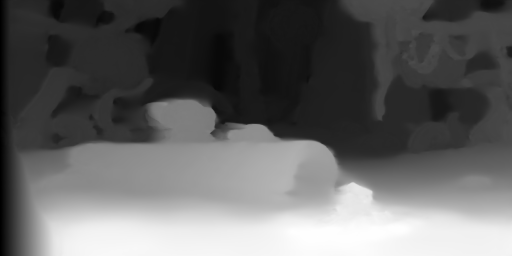} }
\caption{\label{fig:result_mk}\small \textbf{Our qualitative results on Monkaa:} From top to bottom: Left image, reconstructed left image, recovered disparity map. From left to right: frame 11, frame 15, frame 133, frame 143.}
\end{center}
\end{figure}

\begin{figure}[!ht]
\begin{center}
\subfigure{
\includegraphics[width=0.22\linewidth]{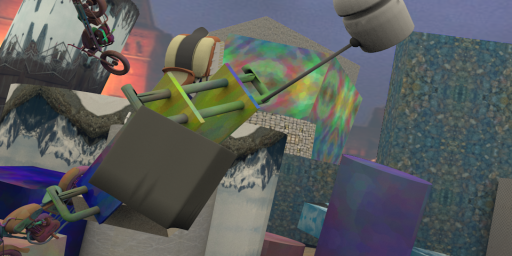} }
\subfigure{
\includegraphics[width=0.22\linewidth]{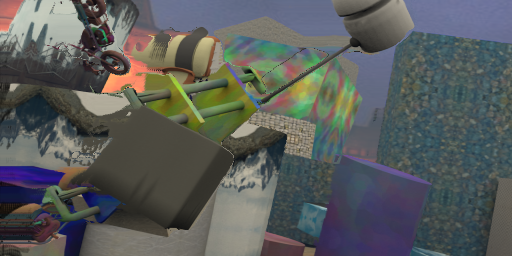} }
\subfigure{
\includegraphics[width=0.22\linewidth]{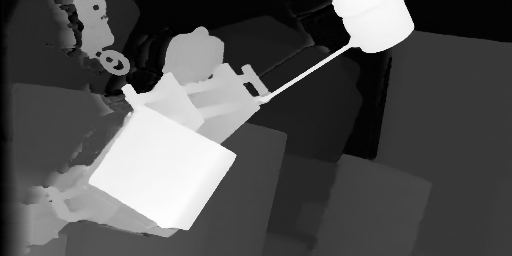} }
\subfigure{
\includegraphics[width=0.22\linewidth]{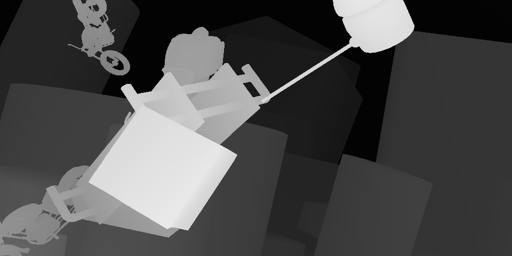} }
\subfigure{
\includegraphics[width=0.22\linewidth]{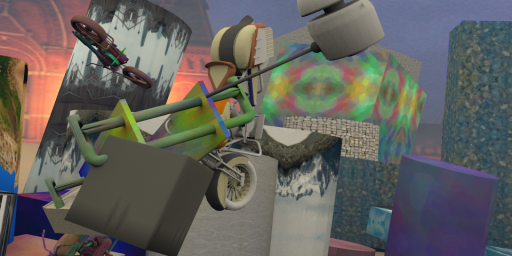} }
\subfigure{
\includegraphics[width=0.22\linewidth]{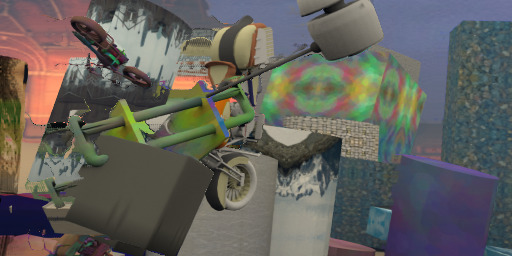} }
\subfigure{
\includegraphics[width=0.22\linewidth]{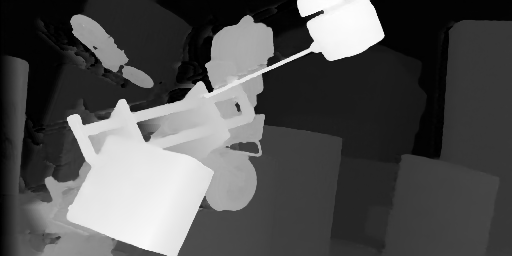} }
\subfigure{
\includegraphics[width=0.22\linewidth]{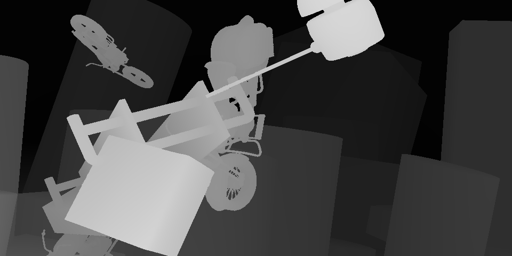} }
\caption{\label{fig:result_cr}\small \textbf{Our qualitative results on FlyingThings3D:} From left to right: Left image, reconstructed left image, estimated disparity map, and ground truth.}
\end{center}
\end{figure}

\textbf{Freiburg Sceneflow Dataset.}
We select two stereo videos from the Monkaa and FlyingThings3D dataset \cite{Mayer2016CVPR} and directly feed them into our network. Qualitative results are shown in Figure-\ref{fig:result_mk} and Figure-\ref{fig:result_cr} correspondingly. Our network produces very accurate disparity maps when compared with the ground truth disparity maps. Furthermore, the reconstructed color images with the estimated disparity map further prove the effectiveness of our model.

\begin{figure}[!ht]
\begin{center}
\subfigure{
\includegraphics[width=0.20\linewidth]{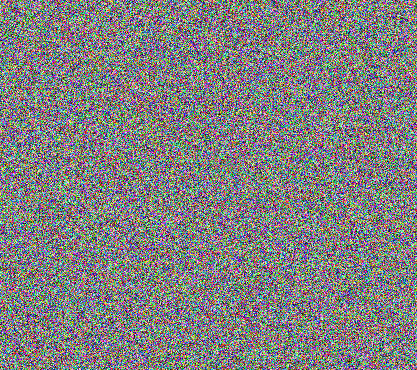} }
\subfigure{
\includegraphics[width=0.20\linewidth]{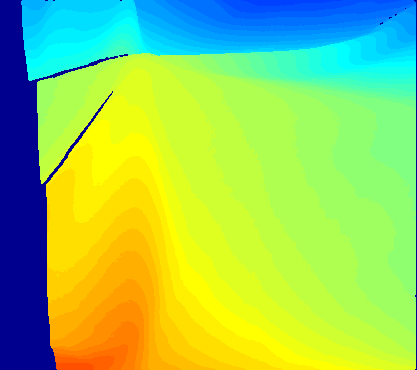} }
\subfigure{
\includegraphics[width=0.20\linewidth]{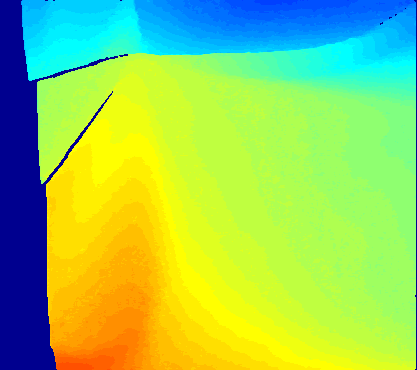} }
\subfigure{
\includegraphics[width=0.20\linewidth]{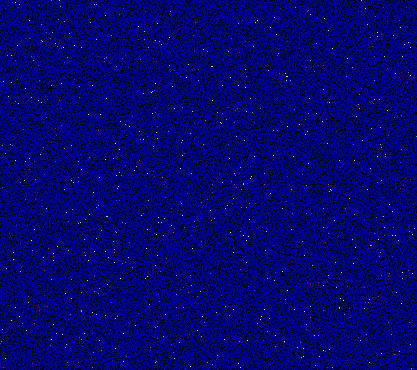} }
\caption{\small \label{fig:result_random} \textbf{Test results on RDS (Random dot stereo) images:}  Left to right: input stereo (left image), ground-truth disparity (color coded), our result, result by Dispnet. Our method successfully recovers the correct disparity map, demonstrating its superior generalibility on unseen images.}
\end{center}
\end{figure}

\textbf{Random dot stereo.}
We test the behavior of our OpenStereoNet on random dot stereo images where there is no semantic content in the images.  Our network works well, however the DispNet fails miserably as shown in Figure-\ref{fig:result_random}.

\section{Conclusions and Discussions}
This paper addresses a practical demand of deploying stereo matching technique to unconstrained real-world environments with previously unseen or unfamiliar ``open-world'' scenarios. We envisage such a stereo matching method that is able to take a continuous live stereo video as input, and automatically predict the corresponding disparity maps.  To this end, this paper has proposed a deep Recurrent Neural Network (RNN) based stereo video matching method--{\em OpenStereoNet}. It consists of a CNN Feature-Net, a Match-Net and two convolutional-LSTM recurrent blocks to learn temporal dynamics in the scene.  We do notice that finding optical flow (or scene flow) between image frames is yet another feasible paradigm to encode and to exploit temporal dynamics existed in a video sequence. However, we argue optical flow itself is a significant research topic in itself, no less challenging than stereo matching, and a comparison between the two approaches deserves to be a valuable future work.

Our OpenStereoNet does not need ground-truth disparity maps for training. In fact, there is even no clear distinction between training and testing as the network is able to learn on-the-fly, and to adapt itself to never-seen-before imageries rapidly.  We have conducted extensive experiments on various datasets in order to validate the effectiveness of our network. Importantly, we have found that our network generalizes well to new scenarios. Evaluated based on absolute performance metrics for stereo, our method outperforms state-of-the-art competing methods by a clear margin.

\noindent
\textbf{Acknowledgements}
Y. Zhong's PhD scholarship is funded by CSIRO Data61. H. Li's work is funded in part by Australia ARC Centre of Excellence for Robotic Vision (CE140100016). Y. Dai is supported in part by National 1000 Young Talents Plan of China, Natural Science Foundation of China (61420106007, 61671387), and ARC grant (DE140100180). The authors are very grateful to NVIDIA's generous gift of GPUs to ANU used in this research.

\bibliographystyle{splncs}
\bibliography{Stereo-Matching-Reference}


\end{document}